\documentclass{Interspeech}

\interspeechcameraready 
\usepackage{soul}
\usepackage{hyperref}
\usepackage{cleveref}
\newcommand\autocite{\cite} 
\newcommand\textcite{\cite}
\newcommand{\ignore}[1]{}

\renewcommand{\paragraph}[1]{\noindent\textbf{#1}\quad}

\title{On the reliability of feature attribution methods for speech classification}
\author[affiliation={1}]{Gaofei}{Shen}
\author[affiliation={1}]{Hosein}{Mohebbi}
\author[affiliation={2}]{Arianna}{Bisazza}
\author[affiliation={1}]{Afra}{Alishahi}
\author[affiliation={1}]{Grzegorz}{Chrupa\l{}a}

\affiliation{}{Tilburg University}{The Netherlands}
\affiliation{}{University of Groningen}{The Netherlands}
\email{\{g.shen, h.mohebbi, a.alishahi\}@tilburguniversity.edu,\\ a.bisazza@rug.nl, grzegorz@chrupala.me}

\usepackage{comment}

\begin{document}

\maketitle

\begin{abstract}
As the capabilities of large-scale pre-trained models evolve, understanding the determinants of their outputs becomes more important. Feature attribution aims to reveal which parts of the input elements contribute the most to model outputs. In speech processing, the unique characteristics of the input signal make the application of feature attribution methods challenging. We study how factors such as input type and aggregation and perturbation timespan impact the reliability of standard feature attribution methods, and how these factors interact with characteristics of each classification task.  We find that standard approaches to feature attribution are generally unreliable when applied to the speech domain, with the exception of word-aligned perturbation methods when applied to word-based classification tasks.
\footnote{Code: https://github.com/techsword/reliability-speech-feat-attr}

\end{abstract}

\section{Introduction}
Large-scale self-supervised models such as wav2vec2 \autocite{baevskiWav2vec20Framework2020} and HuBERT \autocite{hsuHuBERTSelfSupervisedSpeech2021} have shown impressive performance on various downstream speech processing tasks from automatic speech recognition to audio classification.
As transformer models \autocite{vaswaniAttentionAllYou2017} have been increasingly adopted in speech processing, interpretability research for these models has also intensified.

An important research domain within interpretability is~ \textit{feature attribution} which aims to quantify the contribution of different parts of the model input to its output. 
A variety of approaches to feature attribution been studied extensively for the domains of computer vision (CV) and natural language processing (NLP) models. A more limited body of work also exists for the domain of spoken language \autocite{prasadHowAccentsConfound2020,beckerAudioMNISTExploringExplainable2024,wuExplanationsAutomaticSpeech2023,wuCanWeTrust2024,guptaPhonemeDiscretizedSaliency2024,fucciSPESSpectrogramPerturbation2024,manciniInvestigatingEffectivenessExplainability2024,pastorExplainingSpeechClassification2023}.

A key challenge in research on feature attribution is the evaluation of the methods. Following the long-standing practice in research methods and psychometrics, we can distinguish two key concepts used to quantify the quality of a measurement method: \textit{reliability} and \textit{validity}. Validity refers to how well the method measures the quantity of interest. For feature attribution it largely overlaps with the concept of faithfulness: does the attribution highlight the features that in reality are the most important determinant features of the model's output? On the other hand, reliability focuses on the consistency of a measurement, and answers the question: does the measurement give the same answer when repeated under similar conditions? For a measurement to be useful, it needs to score well on both of these dimensions. In this work we focus on evaluating the \textbf{reliability} of commonly used feature attribution methods as applied to speech models. This aspect of evaluation is often neglected in prior research, but it is crucial to ensure that the whole endeavor of attributing model outputs to inputs rests on a solid foundation. Only when methods are shown to be reliable can we then focus on the question of their validity.

We believe this foundational work is especially needed in the speech domain. In contrast to the relatively intuitive saliency maps for CV or token-based attribution scores for NLP, the continuous and high resolution nature of the speech signal means that naive application of basic attribution methods leads to noisy and hard-to-interpret results. The choice of feature attribution methods for a particular task matters, as does the conditions under which the method is applied. In this paper, we focus on four attribution methods, and investigate two main aspects that might affect their reliability: the \textbf{input type} used for the feature attribution analysis, and the choice of the \textbf{attribution granularity}, or the timespan of aggregation or perturbation of input of a given attribution method.

\paragraph{Input type.}
Convolutional neural network (CNN) speech classification models can use either spectrograms and waveforms as input \autocite{beckerAudioMNISTExploringExplainable2024}. Meanwhile self-supervised transformer-based models like wav2vec2 \autocite{baevskiWav2vec20Framework2020} and HuBERT \autocite{hsuHuBERTSelfSupervisedSpeech2021} operate in an end-to-end fashion and use waveform as input with a CNN block serving as the feature extractor. More traditional audio features such as log-Mel spectrogram are still being used by the popular Whisper model \autocite{radfordRobustSpeechRecognition2022}. It is important to note that a spectrogram (time-frequency representation), and raw waveform are two different feature representations of the same \textit{input signal}. We can convert a spectrogram to a waveform and vice versa. Thus to a large extent we can decouple input type from the specific target model.
A more model-specific option is to attribute to the output of the CNN feature extractor block of models like wav2vec2 and HuBERT. Following the common practice in NLP, we call the output of the CNN feature extractor the \textit{CNN embedding}. We can think of the CNN embedding as an even higher-level representation of the input signal than the typical time-frequency representation. Thus, for the most common models of interest we need to decide which input type is the most appropriate to use for feature attribution.

\paragraph{Attribution granularity.}
While both gradient-based and perturbation-based methods have been tested on speech models, the interpretation of the attribution results differs. Gradient-based methods assign a score to every input value. The standard 16kHz sampling rate for wav2vec2 models means there are tens of thousands of attribution values for mere seconds of speech. These scores can then be aggregated over longer time-spans for ease of interpretation, but that happens as a post-processing step. Perturbation-based methods, on the other hand, can be applied in a top-down manner by directly perturbing larger chunks of the input signal and observing the change in the model's output. Thus, the choice of attribution method is coupled with the timespan of aggregation or perturbation. 

In this work we investigate four feature attribution methods and quantify the impact of the choices regarding input type and granularity on their reliability. We apply these methods to speech classification models trained on three different tasks (one of which comprises three related subtasks). In order to quantify reliability, we use feature attribution agreement between pairs of separate training runs trained on the same data and applied to the same test input: we name this reliability score inter-(random)-seed-agreement (ISA). Our experiments show that even though the target classifier models learn the tasks and agree on the vast majority of the inputs, the attribution agreement is generally quite low, and that acceptable levels of reliability are only reached in very few specific conditions. Our findings highlight the inadequacy of standard approaches to feature attribution as applied to the speech domain, and underline the need for the development and careful evaluation of appropriate speech-specific attribution methods.

\section{Related Work}
Feature attribution has evolved along with advances in machine learning, with methods originally developed to visualize salient features in computer vision \autocite{zeilerVisualizingUnderstandingConvolutional2014, simonyanDeepConvolutionalNetworks2014}, and adapted further for natural language processing \cite{Covert2020ExplainingBR, bastings-filippova-2020-elephant, mohebbi-etal-2021-exploring}.
Unlike in speech processing, the limitations of feature attribution methods have been carefully studied in vision and language models from both reliability and validity perspectives \cite{pruthi-etal-2022-evaluating, krishna2024the, Neely2022ASO, Sixt2019WhenEL, Bilodeau2022ImpossibilityTF}. For example, gradient-based methods were demonstrated to be independent of parameters of later layers \cite{Sixt2019WhenEL}, while \cite{Bilodeau2022ImpossibilityTF} argues that complete and linear attribution methods (such as Integrated Gradients) may perform no better than random guessing when identifying how models depend on features. By computing rank correlation, \cite{Neely2022ASO, krishna2024the} show that feature attribution methods (even those within the same family) often disagree in the explanation scores they produce. Unlike these approaches, we do not consider agreement between different attribution methods, but rather focus on the core issue of the reliability of a {\it single method} under {\it repeated measurement}: the consistency of a single attribution method applied to several randomly initialized fine-tuning runs of the same model architecture. Regarding the effect of different configuration details (e.g., aggregation level), our work is related to \cite{bastings-etal-2022-will, chenDynamicMultigranularityAttribution2024} for textual data.

Applications and adaptations of attribution methods in the speech domain has been explored primarily with convolutional neural network (CNN) based models \cite{prasadHowAccentsConfound2020, beckerAudioMNISTExploringExplainable2024,muckenhirnUnderstandingVisualizingRaw2019}, given their architectural similarities with computer vision models. For example, \textcite{beckerAudioMNISTExploringExplainable2024} used Layerwise Relevance Propagation (LRP) to explain a CNN model trained on either waveform or spectrogram representations of audio signals. In contrast to their approach, where models trained on two different types of input data are compared, we examine the effect of different input types within a single model.

For speech classification, several studies have started to examine the validity of feature attribution methods. For example,
\cite{wuCanWeTrust2024} applied LIME \cite{ribeiroWhyShouldTrust2016} to a phoneme recognition task using the TIMIT \cite{garofolojohns.TIMITAcousticPhoneticContinuous1993} dataset, which provides manual labeling and segmentation at the phoneme level. They found that restricting input audio perturbations to a limited window around the phoneme of interest can improve the validity of LIME. Similarly, \textcite{guptaPhonemeDiscretizedSaliency2024,pastorExplainingSpeechClassification2023} showed that discretizing attribution scores through phoneme- and word-level boundaries leads to more interpretable explanations for classification tasks.
Despite these studies on validity, the reliability of feature attribution for speech models remains largely unexplored.

\section{Methods}
For speech classification, pre-trained models are typically paired with a lightweight feedforward neural network as a classification head and fine-tuned using labeled data. During fine-tuning, both the backbone model and classifier adjust their weights to emphasize the most relevant input features and learned representations, maximizing classification accuracy. We therefore assume that models starting from the same pre-trained checkpoint but fine-tuned with different random seeds identify and employ comparable relevant features for a given input, especially when they achieve consistently high accuracy. 
Accordingly, a reliable feature attribution method should consistently show the same pattern in highlighting the most important input features for a given utterance across such models.
\subsection{Classification models}
\begin{table}
    \centering
    \scriptsize
    \begin{tabular}{l c c c c}
        \hline
        Task           & Accuracy & Overall Fleiss' $\kappa$ & Error Fleiss' $\kappa$ \\
        \hline
        Gender ID      & 0.999    & 0.999                    & 0.356                  \\
        Speaker ID     & 0.990    & 0.983                    & 0.677                  \\
        Intent Class.  & 0.998    & --                       & --                     \\
        \quad Action   & 0.997    & 0.997                    & 0.520                  \\
        \quad Object   & 0.999    & 0.999                    & 0.602                  \\
        \quad Location & 0.999    & 0.999                    & 0.591                  \\
        \hline
    \end{tabular}
    \caption{Agreement in model performance across different runs, measured using three metrics: Accuracy, Overall and Error Fleiss' $\kappa$ on the test sets.}

    \label{tab:perf-error-analysis}
\end{table}

In our experiments, we use \verb|wav2vec2-base|\footnote{https://huggingface.co/facebook/wav2vec2-base} model and fine-tune it with nine different seeds for three different speech classification tasks: gender and speaker identification (Gender ID and Speaker ID respectively), and intent classification (IC). We expect the IC task to rely on mostly on the presence of specific lexical items, while Gender~ID and Speaker~ID should rely mostly on lower-level acoustic features.

For Gender ID and Speaker ID, we use a subset of the Common Voice dataset \cite{ardilaCommonVoiceMassivelyMultilingual2020}. We select 40 speakers with self-reported gender labels of masculine or feminine with 301 utterance for each speaker totalling 12,040 utterances. A stratified 80:20 train-test split was also applied before model fine-tuning. For intent classification we use the Fluent Speech Commands dataset \cite{lugoschSpeechModelPretraining2019} with the provided train-test split. We resample the waveforms in both datasets to 16kHz. The IC task comprises three related classification subtasks: Action, Object and Location. We use a separate classification head for each subtask. 

During fine-tuning, we freeze the CNN feature extractor and the projection layers and only update weights of the transformer network and the final classification heads. This makes sure models with different seeds receive exactly the same input and  allows us to have a fixed \textit{CNN embedding} across different models investigated in this paper.

In order to lend further credibility to our assumption about the equivalency of models, we first evaluate the classification accuracy as well classification agreement of the different fine-tuning runs, shown in \Cref{tab:perf-error-analysis}. If models' architecture as well as behavior is similar, we have more reason to be believe that their internal computations are also equivalent.

To measure the agreement between the model runs initialized with different random seeds, we report two separate version of  Fleiss' $\kappa$. The overall Fleiss' $\kappa$ measures the agreement between all runs on the complete test set. The Error Fleiss' $\kappa$ measures the agreement between model decisions on the subset of the test data where at least one error was made. Fleiss' $\kappa=0$ if the agreement between runs is due to chance, and $\kappa=1$ if the runs agree completely.

We can see that the classification accuracy is near perfect for all tasks and overall agreement is also very high.  Agreement for the small subset of data points where an error was made is moderate. Given small percentage of overall errors, we believe this means the behavior of the models we investigate in this paper are sufficiently similar.

\subsection{ISA reliability metric}
To measure the reliability of feature attribution methods, we examine the consistency of their scores derived from models fine-tuned with different seeds.
Specifically, we calculate the inter-seed agreement (ISA) metric based on a dynamic \textit{top-p} metric.
\begin{equation}
    \text{ISA} = \frac{1}{N} \sum_{n=1}^{N} \frac{|\text{top-p}(A_i)_n \cap \text{top-p}(A_j)_n| }{|\text{top-p}(A_i)_n| }
\end{equation}
Here $p$ is the percentage of the top indices of the attribution scores we are interested in. $A_i$ and $A_j$ are the attribution scores for the $i$-th and $j$-th model run. $N$ is the number of samples in the dataset. The top-p function returns the top $p$ percentage of indices of the attribution scores. The value of $p$ is fixed at 20\%.
The intersection of the top-p indices is calculated for each sample in the dataset. The ISA score is the average of the pairwise percentage of shared indices of the top-p attribution scores for all combinations of random seed pairs. The higher the ISA score, the more the attribution scores of the models agree with each other. A baseline ISA can be obtained by randomly shuffling the attribution scores for each sample before calculating the ISA score.

We use the Captum library \cite{kokhlikyanCaptumUnifiedGeneric2020} to calculate feature attribution scores for the fine-tuned models on their respective datasets. We test two gradient-based methods: Saliency and Integrated Gradients (IG); and two perturbation-based methods: LIME and Feature Ablation (FA). For the perturbation-based methods, we use a feature mask to group waveform input into 10ms spans: this is done due to computational constraints as a highest tractable resolution for perturbing the waveform.

\subsection{Feature attribution conditions}

\paragraph{Input feature types.} We calculate the ISA metric for each of our three input feature types: waveform, spectrogram, and CNN embeddings. In order to enable attribution to the spectrogram for self-supervised models like wav2vec2 and HuBERT (which use raw waveform as their native input), we follow \autocite{manciniInvestigatingEffectivenessExplainability2024} by prepending an inverse short-term Fourier transformation (ISTFT) to the model. We use a hop length of 320 for the STFT and ISTFT transformations to keep the time resolution at 20ms to be consistent with the wav2vec2 model feature extractor.

For Integrated Gradients, LIME and Feature Ablation, a baseline input is needed. We use silence as our baseline: for the waveform input type we use the silence waveform directly, while for the other two input types we convert the silence waveform into the corresponding spectrogram or CNN embedding first. All of the attribution methods tested return both positive and negative values: we do not do any additional processing to the attribution score before aggregation.

\paragraph{Granularity of aggregation.} We test three granularities of aggregation: no aggregation, frame-level aggregation, and word-level aggregation. For frame-level aggregation, we sum the attribution scores for raw waveform input at 20ms intervals; spectrogram and CNN embedding input gets summed for every frame.
For word-level aggregation, we aggregate the attribution scores at the word level using forced-alignment time stamps. We use Montreal Forced Aligner \autocite{mcauliffeMontrealForcedAligner2017} to align our datasets with the provided transcriptions. To take varying word lengths into account, we mean-pool the attribution scores for each word. We also discard the non-word segments in the alignment.

\paragraph{Granularity of perturbation.}
For the perturbation methods only, the alternative to aggregation is to directly perturb specific timespans of the input. We test the effect of directly perturbing word-level segments of the input, based on the same forced-alignment time stamps as above.

\section{Results}

We organize the results into groups of comparisons. Within each group, we present the effects of the varied conditions in applying feature attribution methods. We visualize the central tendency (median) of the inter-seed agreement (ISA) scores as well as the spread around it via boxplots. An individual boxplot displays a set of 36 pairwise comparisons. We then visualize the median baseline score of randomly shuffled attribution scores for all pairwise comparisons via the dashed red lines.

\subsection{Effects of input feature types}

\begin{figure}
    \centering
    \includegraphics[width=0.47\textwidth]{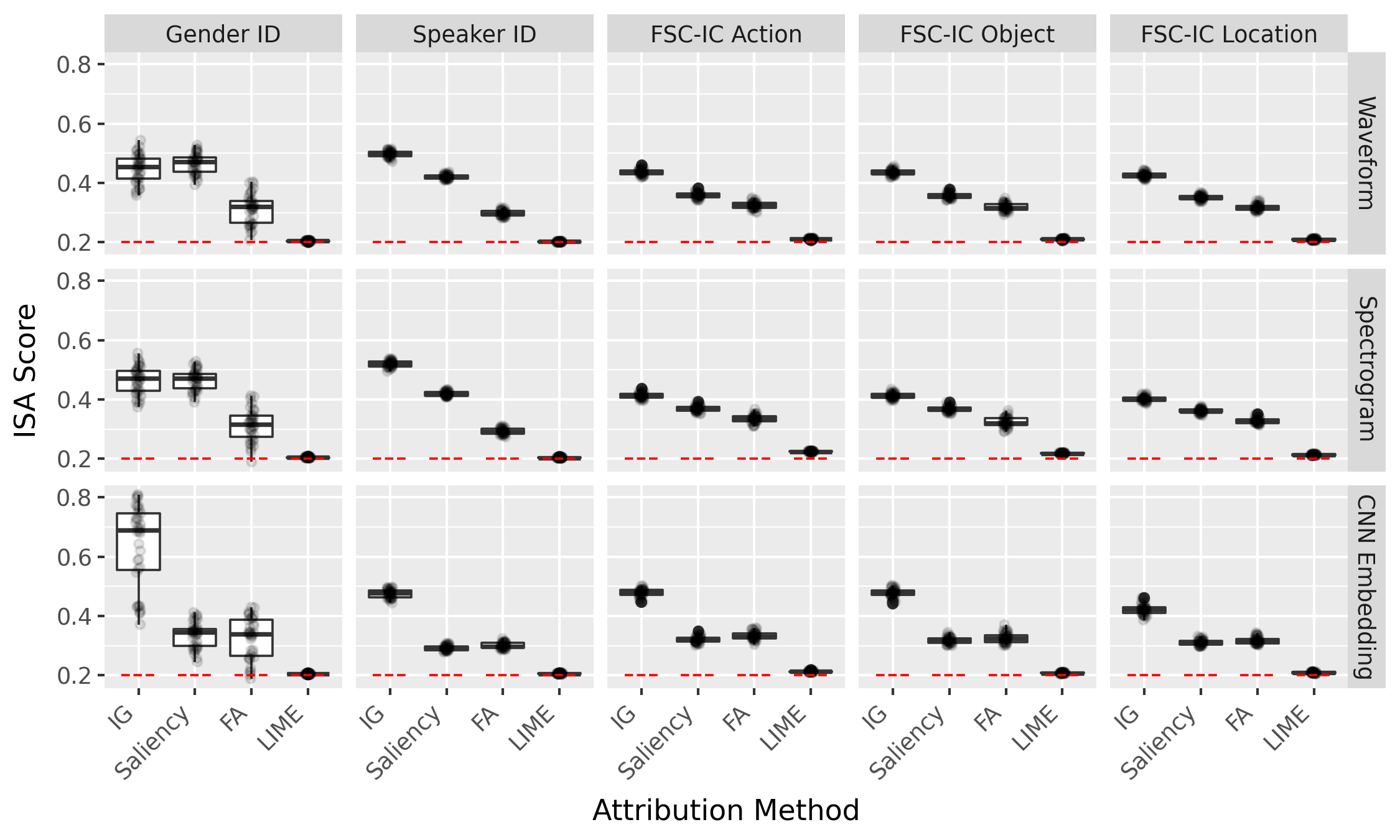}
    \caption{Distributions of ISA scores without aggregation. The rows indicate different input feature types, the columns are different tasks. Within each panel, each boxplot shows results from different attribution methods and the y-axis is the ISA score. The red dotted line indicates the randomly shuffled baseline. IG: Integrated Gradients, FA: Feature Ablation.}
    \label{fig:input_types}
\end{figure}

\Cref{fig:input_types} shows the ISA scores for attribution scores for the no-aggregation condition.
For most combinations of method and input type the ISA scores are low to moderate (below 0.6). The exception is the task of Gender ID for the embedding input type and Integrated Gradients method, which shows a median of around 0.7 but with a wide spread around it.
We also note that the effect of attribution method is in general larger than that of the input type. Notably, in most cases LIME shows very low reliability, barely above the baseline. This indicates that LIME is not a suitable attribution method in a high-time resolution setting.
At the same time, Integrated Gradients generally shows the highest reliability.

\subsection{Effects of granularity of aggregation}
\begin{figure}
    \centering
    \includegraphics[width=0.47\textwidth]{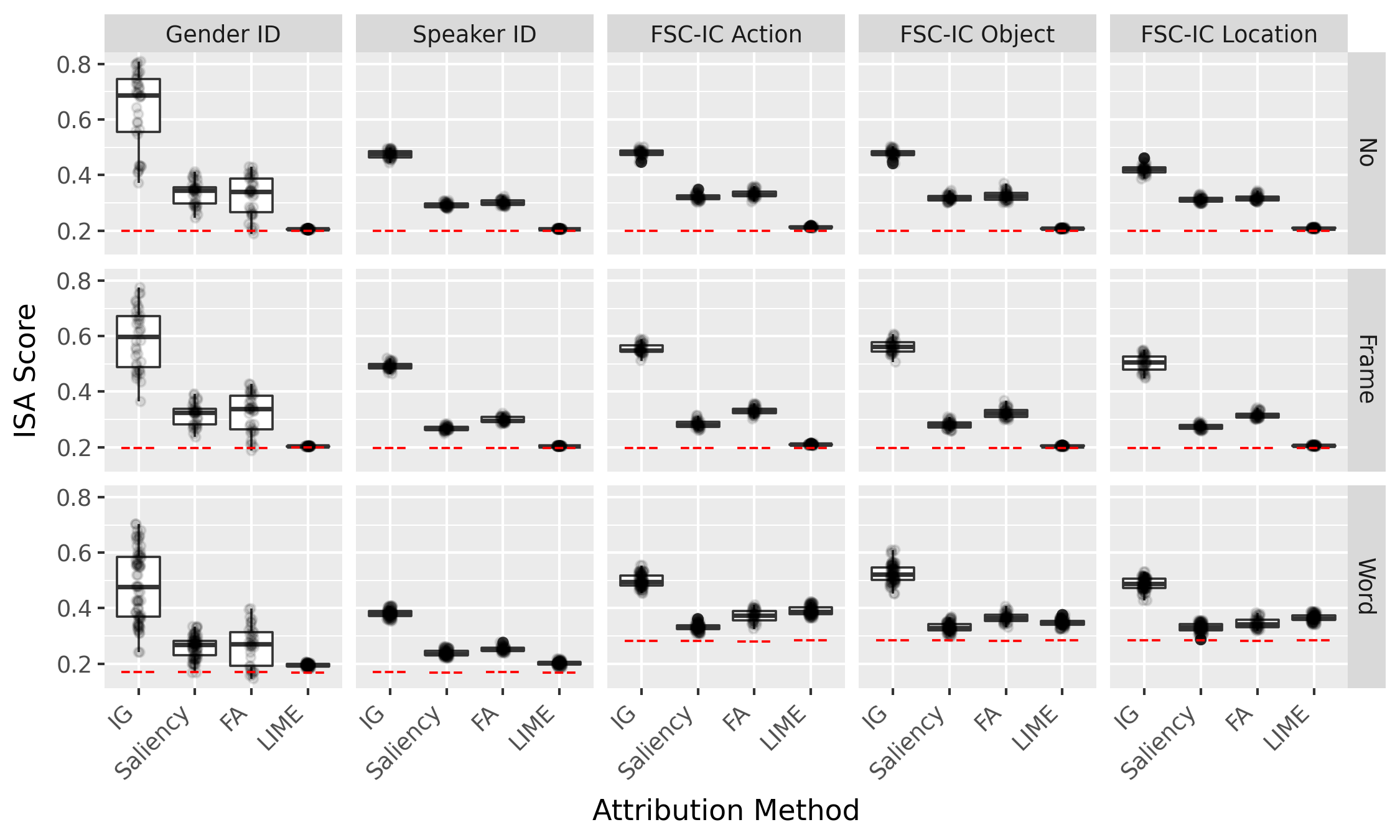}
    \caption{
        Distributions of ISA scores for the CNN embedding input type, at different levels of aggregation. The rows are levels of granularity of aggregation, the columns are different tasks. Within each panel, each boxplot shows results from different attribution methods and the y-axis is the ISA score. The red dotted line indicates the randomly shuffled baseline. IG: Integrated Gradients, FA: Feature Ablation.}
    \label{fig:aggregation_level}
\end{figure}

To highlight the impact of granularity of aggregation, we plot the ISA scores of only the CNN embedding input type at various levels of aggregation in \cref{fig:aggregation_level}.
We can see that different aggregation levels do not alter the general reliability patterns we saw in \cref{fig:input_types}.
As before, we generally see the best reliability for Integrated Gradients. For word-level attribution on the IC tasks we would expect a higher level of agreement, and yet we see that ISA scores do not surpass 0.6 for any of the condition combinations. This is surprising, as intuitively one would expect that aggregation should smooth out small variations in attribution scores and increase agreement.

\subsection{Perturbations on the word level}

\begin{figure}
    \centering
    \includegraphics[width=0.47\textwidth]{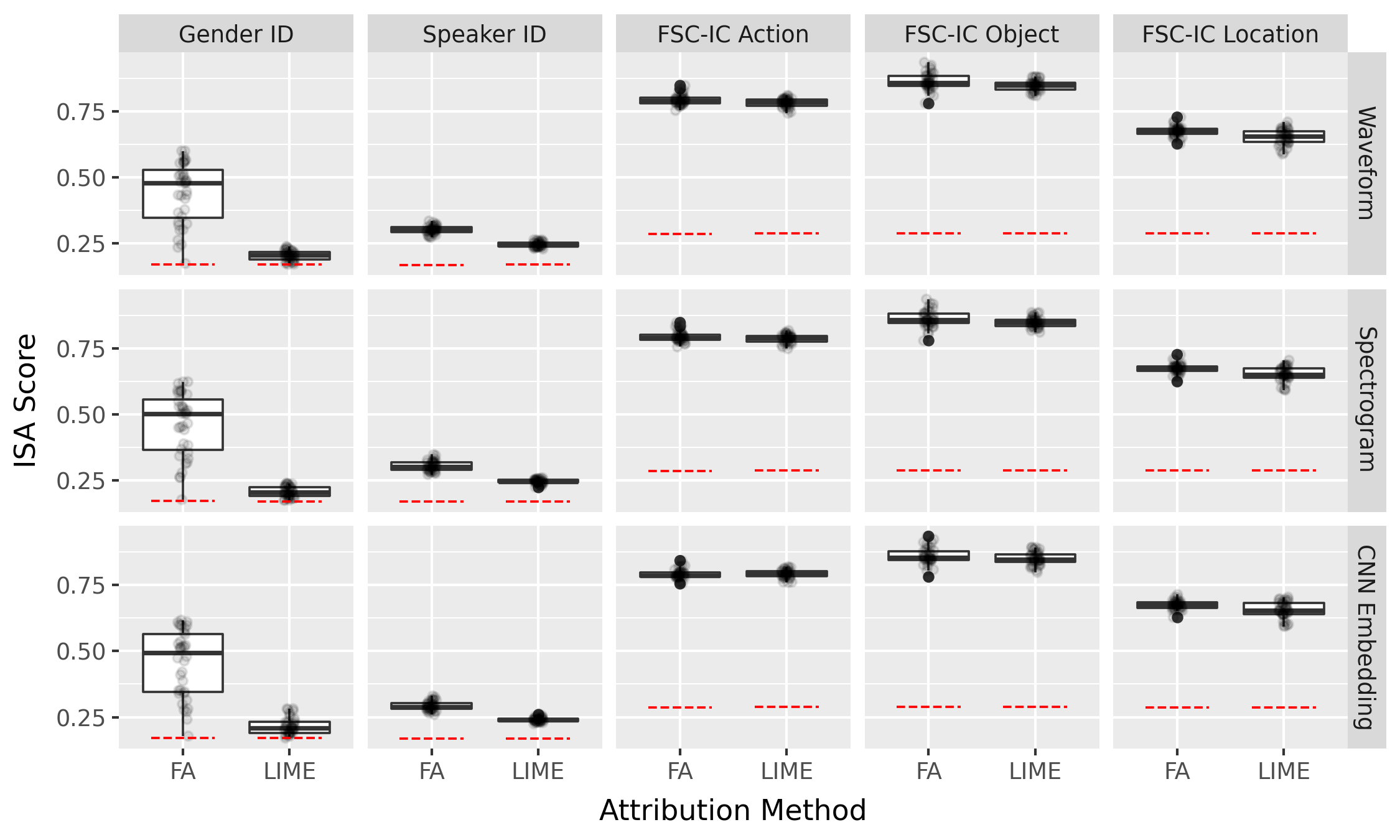}
    \caption{
        Distributions of ISA scores with perturbation operating directly on  word-aligned segments. The rows indicate different input feature types, the columns are different tasks. Within each panel, each boxplot report results from different attribution methods and the y-axis is the ISA score. The red dotted line indicates the randomly shuffled baseline.  FA: Feature Ablation.}
    \label{fig:word_perturbation}
\end{figure}


Lastly, we also evaluate the reliability of perturbation methods applied to word-aligned input segments, in a setting analogous to that described in \cite{pastorExplainingSpeechClassification2023}. \Cref{fig:word_perturbation} shows ISA scores for attribution scores generated with perturbation-based methods operating directly on the word level.
Here we see the Intent Classification tasks showing much higher ISA than in \cref{fig:aggregation_level}. The reliability of perturbation-based attribution methods on word-based tasks is much higher when perturbing word-aligned than aggregating scores obtained from high-resolution features.
We also observe that one of the IC tasks, Location, shows lower reliability than the other two. This perhaps reflects the degree to which these different subtasks can rely on redundant lexical cues. 

Interestingly, we see less variations in the score pattern across different input feature types in \cref{fig:word_perturbation} than in \cref{fig:input_types}. This shows that perturbations done on the word-level granularity are less sensitive to the differences between different input feature types.

\section{Discussion \& Conclusion}

Our findings show that the naive application of standard feature attribution methods to speech classification models generally leads to poor reliability. When attributing to high-resolution input, regardless of specific input types such as waveform, spectrogram or embeddings, even the most reliable of our methods, Integrated Gradients, does not surpass 50\% inter-seed agreement for most tasks. Simply aggregating these scores does not improve reliability. The likely underlying issue is that the gradients or perturbation effect of such high resolution and highly correlated redundant features are very small and noisy.

Only in the case of directly perturbing word-aligned segments of the input, and only for the intent classification subtasks, do we see acceptable reliabilities.
The likely explanation is that classification decisions for these tasks rely on specific words in the utterance, and that directly perturbing those specific words and only those words affects model output and thus attribution scores. There is thus little scope for models to disagree. On the other hand, tasks such as Gender ID and Speaker ID are unlikely to rely on specific words, and models may use redundant clues distributed over the whole utterance: thus for these tasks we do not find a consistently reliable combination of method and input-type to attribute to. Our findings suggest that in order for standard atrribution methods to be applicable, the target speech classification task needs to be similar in nature to an equivalent text-based task, where token-based attributions are standard. Ideally, however, we would like feature attribution to be more wide applicable across varied speech tasks.

\subsection{Limitations and future directions}
While the scope of this paper is limited to assessing reliability, the methods found to be reliable will also need to be evaluated for validity. While previous works have used cross-method agreement as a proxy for validity, we believe that a more direct measure of alignment with the target model will be needed.
Our work focused on reliability for attribution to features along the time dimension. The other important axis for audio data is the frequency domain: for certain tasks it may be more useful to attribute to frequency-based features. Additionally, for audio data generally, and for speech specifically, certain high-level features such as loudness, pitch-contours, or specific aspects of timbre may also be interesting targets for attribution: for these cases standard attribution techniques such as those evaluated here are not directly applicable. Ultimately, spoken language may be sufficiently different from images or text data that only feature attribution techniques tailored to the speech domain will prove reliable enough to be useful.

\section{Acknowledgements}
This publication is part of the project \textit{InDeep: Interpreting Deep Learning Models for Text and Sound} (with project number NWA.1292.19.399) of the National Research Agenda (NWA-ORC) program.

\bibliographystyle{IEEEtran}
\bibliography{custom.bib}

\end{document}